\documentclass{ifacconf}

\usepackage{epsfig}
\usepackage{amsmath}
\usepackage{hhline}
\usepackage{multirow}
\usepackage{enumerate}
\usepackage{float}
\usepackage{graphicx,url}
\usepackage{natbib}

\begin{document}
\begin{frontmatter}

\title{Backstepping Control of Muscle Driven Systems with Redundancy Resolution}


\author[First]{Humberto De las Casas} 
\author[First]{Hanz Richter}

\address[First]{Mechanical Engineering Department, Cleveland State University, Cleveland, OH 44115 USA.} 

\begin{abstract}
Due to the several applications on Human-machine interaction (HMI), this area of research has become one of the most popular in recent years. This is the case for instance of advanced training machines, robots for rehabilitation, robotic surgeries and prosthesis. In order to ensure desirable performances, simulations are recommended before real-time experiments. These simulations have not been a problem in HMI on the side of the machine. However, the lack of controllers for human dynamic models suggests the existence of a gap for performing simulations for the human side. This paper offers to fulfill the previous gap by introducing a novel method based on a feedback controller for the dynamics of muscle-driven systems. The approach has been developed for trajectory tracking of systems with redundancy muscle resolution. To illustrate the validation of the method, a shoulder model actuated by a group of eight linkages, eight muscles and three degrees of freedom was used. The controller objective is to move the arm from a static position to another one through muscular activation. The results on this paper show the achievement of the arm movement, musculoskeletal dynamics and muscle activations.
\end{abstract}
\end{frontmatter}
\section{Introduction}
\noindent Research in human-machine interaction (HMI) has received a lot of attention in recent years. The reason that seems to indicate this fact is the several applications with potential opportunities in the areas of development. For instance, human training has shown enhancement through the use of robotic training machines capable of providing variable resistance based on the user requirements \citep{MIPAPER1} \citep{MIPAPER2}. Furthermore, rehabilitation processes have been proved to be very efficient with the use of machines designed to help people with disabilities to recover their motor skills \citep{MOT2} \citep{MOT3}. Besides, several improvements in the quality of walking have been reported for people with amputees by using prosthesis with energy regeneration \citep{POYA1} \citep{POYA2}. And, in the same way, the quality and precision of surgery, specially at small scales, has been heightened through the use of robots and teleoperation surgeries \citep{Robot_Surgery} \citep{Bianchini2019}. 

\noindent The HMI in the areas of advanced training and rehabilitation have been the main motivation for the development of this work. Robotic machines provide an invaluable contribution to these areas by combining exercise physiology with technology \citep{MOTOR} \citep{HUM_THE}. Moreover, they have the capacity to produce workloads even in lack of gravity showing potential applications for instance in microgravity. The efficiency of the human training is highly important, specially in the space, where it plays a crucial role. Humans operating in lack of gravity for long time periods are exposed to the loss of muscle mass and bone density \citep{NASA_2}. However, robotic machines promise to be beneficial to diminish these detrimental effects \citep{ECC} \citep{ECC2}. 

\noindent Real-time experiments have always played a key role in HMI developments \citep{HMI_SIM}. However, a successful experiment is usually result of a good mathematical model and simulation. Consequently, in order to ensure a desirable performance, simulation tests should be carried out before any real-time experiments. Simulation tests have not been a problem in HMI on the side of the machine. Nonetheless, the lack of accurate human models suggests the existence of a gap. To achieve an accurate model simulation or to adapt a model previously developed can be challenge but highly important \citep{Sim_J1}. HMI simulation models for English conversations between people and a humanoid robot have been reported \citep{HMI_Lang}. The simulation model was developed to help old people with comprehending problems. Another human model for people with abnormal hip stress distribution has been reported \citep{HMI_Hip}. Surgeries are extremely recommended to avoid worsening the disease. However, the optimal surgery procedure can be an arduous process. A simplified human model was the solution to deal with the problem and to support the preoperative planning. And alternatively, this novel method based on a feedback controller of the dynamics of a muscle-driven system is introduced to be used on human training and rehabilitation simulations with muscle activation estimations. This controller for muscle-driven system models makes possible to not only simulate the environment of interaction between a human and a robot, but also to make use of the state estimations as a feedback to regulate the robot behavior.

\noindent The presented method has been developed to control muscle-driven systems. It is based on the backstepping theory for position regulation. This work was originally built on a framework for a lower body muscle-driven \citep{BACKSTEP_2D}. In order to validate the approach, a shoulder model composed by a group of eight linkages, eight muscles and 3 degrees of freedom (DOFs) was used as an application example. The shoulder was conceived as a single ball and socket joint. Therefore, the three DOFs are allowed in a single point. The proposed method shows its ability to control the position and orientation of the shoulder muscle-driven system through muscular activation with redundancy resolution through least squares. 

\noindent The remainder of the paper is organized as follows: Section \ref{S_Dynamics} derives the dynamics of a general muscle-driven system including the linkage and muscle dynamics. Section \ref{S_Control} develops the feedback controller for any muscle driven-system. Section \ref{S_Shoulder} provides the shoulder model dynamics used as the application example. Section \ref{S_Results} shows and discussed the results that validate the developed method. And finally, in Section \ref{S_Future} the conclusions are presented and the future work is stated.

\section{Muscle Driven Dynamics}\label{S_Dynamics}
\noindent The controller proposed in this work makes use of the mathematical model representation of the muscle driven system. This mathematical model is composed by two submodels. The first submodel represents the actuated linkage dynamics.

\subsection{Actuated Linkage Dynamics}
\noindent The linkage dynamics \citep{ROBOT_MODEL} given in joint coordinates are derived as:
\begin{equation}
\label{DE1}
D(q)\ddot{q}+C(q,\dot{q})\dot{q}+g(q)=\tau_{m},
\end{equation}
where $q$ is a vector of joint displacements, $D(q)$ is the inertia matrix, $C(q,\dot{q})$ is the centripetal and Coriolis effects, $g(q)$ is the gravity vector, and $\tau_{m}$ is the control torque produced by the muscle torques. 

\subsection{Muscle Dynamics}
\begin{figure}[ht]
\begin{center}
\includegraphics[width=3.34in]{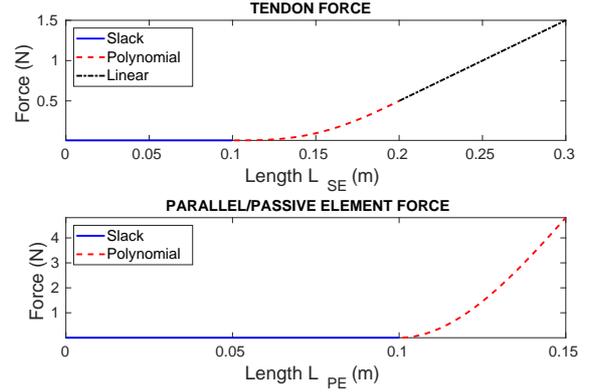}
\caption{Muscle force in function of the muscle elements length.}
\label{fig_F_Tendon}
\end{center}
\end{figure}

\noindent The second submodel represents the muscle dynamics. The linkage dynamics \ref{DE1} are controlled by the muscle torques $\tau_{m}$. These torques are the product of the moment arm and the linear force of the muscle acting on the center of rotation of the joint \citep{MOMENTARM}. The torques are computed as follows:
\begin{equation}
\label{DE2}
\tau_{m}=\overrightarrow{d_m} \times \overrightarrow{F_m},
\end{equation}
where $d_m$ and $F_m$ are the moment arm and the muscle force vectors. The muscle force vector is derived as:
\begin{equation}
\label{DE3}
\overrightarrow{F_m}=\overrightarrow{M_u} \big(\Phi\big),
\end{equation}
where $\overrightarrow{M_u}$ is the unit vector representing the orientation of the muscle, and $\Phi$ is the scalar value of the force along the muscle. The force on the muscle is calculated in function of the length of the serial element or the parallel/passive element according with the muscle representation of the Hill-type model (see Fig. \ref{fig_F_Tendon}).

\subsubsection{Hill-type Muscle Model}
\begin{figure}[ht]
\begin{center}
\includegraphics[width=2.4in]{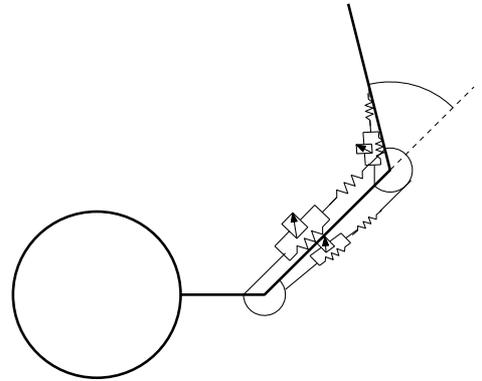}
\caption{Muscles represented as a mechanical system.}
\label{fig_HILL2}
\end{center}
\end{figure}
\begin{figure}[ht]
\begin{center}
\includegraphics[width=2.4in]{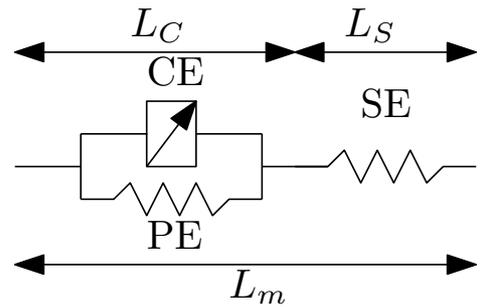}
\caption{Hill-type muscle model.}
\label{fig_HILL}
\end{center}
\end{figure}

\noindent The Hill-type muscle model conceives the musculoskeletal system as a mechanical system (see Fig. \ref{fig_HILL2}). The hill-type muscle model relates the tension with the velocity (regarding the internal thermodynamics) \citep{HILL_MODEL}. The Hill-type muscle model (see Fig. \ref{fig_HILL}) consists of following three elements:

\begin{itemize}
\item A parallel elastic element (PE): Representing the collagen tissue.
\item A contractile element (CE): Representing the contractile properties.
\item A serial element (SE): Representing the tendinous tissue.
\end{itemize}

The length of the muscle $L_{m}$ is computed as:
\begin{equation}
\label{MUSC1}
L_{m}=L_{CE}+L_{SE},
\end{equation}
where the $L_{CE}$ and $L_{SE}$ are the length of the contractile and serial element respectively. Besides, the rate of change of the contractile element length represents the activation on the muscle ($-\dot{L}_{CE}=u$). This activation $u$ served as the control input to the muscle control. Consequently, from Eqn. \ref{MUSC1}, the speed of contraction for the serial element $\dot{L}_{SE}$ is derived as:
\begin{equation}
\label{MUSC2}
\dot{L}_{SE}=\dot{L}_{m}+u
\end{equation}

\section{Feedback Controller}\label{S_Control}
\noindent The feedback controller proposed in this work is based on the backstepping theory \citep{backstep_theo}.

\subsection{Feedback Control by Backstepping}
\noindent From Eqn. \ref{DE1}, a synthetic control is defined as the control torque ($\zeta \overset{\Delta}{=} \tau_m$). And a feedback law $\Psi(e)$ for tracking control is introduced. The variable $e$ denotes the tracking error as
\begin{equation}
\label{C1}
e=\begin{bmatrix}
\tilde{q}\\\dot{\tilde{q}}
\end{bmatrix}=\begin{bmatrix}
q-q_{des}\\ \dot{q}-\dot{q}_{des}
\end{bmatrix}
\end{equation}
\noindent The feedback law $\Psi(e)$ is built on the basis of inverse dynamics \citep{IK} as follows:
\begin{equation}
\label{C2}
Da+C(q,\dot{q})\dot{q}+g(q),
\end{equation}
where $a$ is the synthetic acceleration defined as $a=\ddot{q}_{des}-Kd\tilde \dot{q} - Kp\tilde{q}$ with $Kd$ and $Kp$ as diagonal matrices of positive gains.

\noindent A control error $w$ is defined as the difference between the synthetic control and the synthetic feedback law ($w=\zeta-\Psi$).

\noindent A perfect tracking ($e=[0,0]^T$) and a decreasing positive definite Lyapunov function are assumed when the synthetic control and the synthetic feedback law are equal ($w=0$).

\noindent When the control error is not zero ($w\neq0$), the error dynamics can be found by rewriting the dynamic equation for the actuated linkage system (Eqn. \ref{DE1}) as
\begin{equation}
\label{C3}
M(q)\ddot{q}+C(q,\dot{q})\dot{q}+g_v(q)=\zeta=\Psi+w,
\end{equation}
and by substitution of the inverse dynamics from Eqn. \ref{C2}, the control error is defined as:
\begin{equation}
\label{C4}
w=M\ddot{\tilde{q}}+MK_d\dot{\tilde{q}}+MK_p\tilde{q}
\end{equation}
Thereby, the error dynamics ($e$) are designated as
\begin{equation}
\label{C5}
\dot{e}=Ae+Bw,
\end{equation}
where
\begin{equation}
\label{C6}
A=\begin{bmatrix}
  0&I\\
  -K_p&-K_d\\
\end{bmatrix}
\quad \quad \text{and} \quad \quad
B=\begin{bmatrix}
 \quad 0\quad\\\quad M^{-1}\quad
\end{bmatrix}.
\end{equation}

\subsection{Lyapunov Function}
\noindent The Lyapunov function is derived as
\begin{equation}
\label{LYAP1}
V=e^TPe+w^TRw,
\end{equation}
where $P=P^T>0$ and $R=R^T>0$ to ensure the positive definiteness.

\noindent The time derivative of the Eqn \ref{LYAP1} is computed using Eqn. \ref{C5} as follows
\begin{equation}
\label{LYAP3}
\dot{V}=-e^TQe+2w^T(B^TPe+R(\dot{\zeta}-\dot{\Psi}))
\end{equation}
The negative-definiteness of $\dot{V}$ (Eqn. \ref{LYAP3}) is enforced by making
\begin{equation}
\label{LYAP4}
2w^T(B^TPe+D(\dot{\zeta}-\dot{\Psi})=-\Gamma w,
\end{equation}
where $\Gamma=\Gamma^T>0$. And by solving Eqn. \ref{LYAP4} for the time derivative of the synthetic control $\dot \zeta$
\begin{equation}
\label{LYAP5}
\dot\zeta=\dot\Psi-R^{-1}(\Gamma w+B^TPe)
\end{equation}

\subsection{Control Input}
From Eqn. \ref{MUSC2}, the control input (in matrix form) is represented as
\begin{equation}
\label{CI1}
U_i=\dot{L_i}-\dot{S_i},
\end{equation}
where $S_i$ (the vector of series element lengths) is part of the derivative of the synthetic control ($\zeta$)
\begin{equation}
\label{CI2}
\dot\zeta=\frac{\partial \tau_i}{\partial q_i}\dot{q_i}+\bigg[\frac{\partial \tau_i}{\partial S_i}\bigg]^T\dot{S_i}
\end{equation}
Using the value calculated for $\zeta$ in Eqn. \ref{LYAP5} and replacing Eqn. \ref{CI1} in Eqn. \ref{CI2}, the control input is calculated as
\begin{equation}
\label{CI3}
U_i=\Bigg(\bigg[\frac{\partial \tau_i}{\partial S_i}\bigg]^T\Bigg)^+\Bigg(\dot\zeta_i-\frac{\partial \tau_i}{\partial q_i}\dot{q_i}-\bigg[\frac{\partial \tau_i}{\partial S_i}\bigg]^T\dot{L_i}\Bigg)
\end{equation}
Where $N^+$ represents the pseudoinverse of $N$.

\section{Application Example}\label{S_Shoulder}
\begin{figure}[ht]
\begin{center}
\includegraphics[width=3.34in]{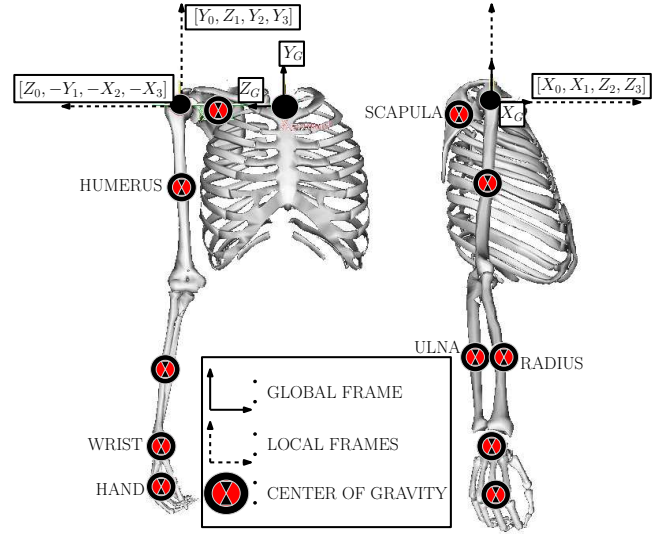}
\caption{Upper arm skeletal system and frames.}
\label{fig_Frames}
\end{center}
\end{figure}

\noindent The approach is validated by using a shoulder musculoskeletal system. The system was modeled as a group of eight linkages, eight muscles and 3 DOFs (see Fig. \ref{fig_Frames}). The model dynamics are available for free download at \citep{Humberto_GITHUB}.

\begin{figure}[ht]
\begin{center}
\includegraphics[width=3.34in]{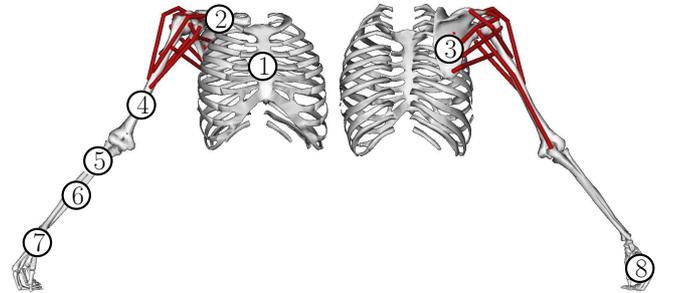}
\caption{Simulated bodies (anterior and posterior musculoskeletal system view).}
\label{fig_Bodies}
\end{center}
\end{figure}

\begin{figure}[ht]
\begin{center}
\includegraphics[width=3.34in]{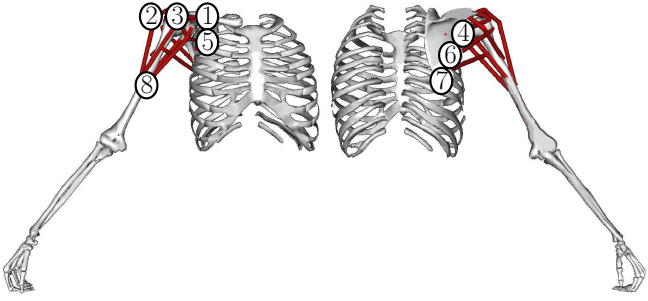}
\caption{Simulated muscles (anterior and posterior musculoskeletal system view).}
\label{fig_Muscles}
\end{center}
\end{figure}

\begin{table}[ht]
\begin{center}
\caption{Linkages of the musculoskeletal system.}
\label{table_bones}
\begin{tabular}{l l l l}
\hline
N$^o$&Linkage & Position (relative to)& Attached Muscles \\
\hline
1&Ground&Fixed&0\\
2&Clavicle&Fixed (Ground)&1\\
3&Scapula&Fixed (Clavicle)&9\\
4&Humerus&Mobile (Scapula)&9\\
5&Ulna&Fixed (Humerus)&1\\
6&Radius&Fixed (Ulna)&0\\
7&Wrist&Fixed (Radius)&0\\
8&Hand&Fixed (Wrist)&0\\
\hline
\end{tabular}
\end{center}
\end{table}

\noindent The shoulder model (including muscles and bones length, mass, inertia and center of mass) was developed based on the real parameters of a person with 1.8 meters of tall and a mass of 75.16 kg. The data was extracted using the OPENSIM software \citep{OPENSIM}. The OPENSIM model used was originally developed for the complete upper body \citep{OPENSIM_MODEL}. 

\noindent A total of eight linkages and eight muscles were selected to fit the model (see Fig. \ref{fig_Bodies} and Fig. \ref{fig_Muscles}). Three of the eight linkages are fixed to ground (see Table. \ref{table_bones}). The selection of the muscles has been done based on those which have contribution to the shoulder movement. Due to the large size of the deltoid, the muscle was modeled as two individual smaller muscles (Deltoid-1 and Deltoid-2). The muscle architecture is listed on Table. \ref{table_muscles}.

\begin{table}[ht]
\caption{Muscles in the musculoskeletal system.}
\begin{center}
\label{table_muscles}
\begin{tabular}{l l l l}
\hline
N$^o$&Muscle & Attachment-1& Attachment-2 \\
\hline
1&Deltoid-1&Humerus&Clavicle\\
2&Deltoid-2&Humerus&Scapula\\
3&Supraspinatus&Humerus&Scapula\\
4&Infraspinatus&Humerus&Scapula\\
5&Subscapularis&Humerus&Scapula\\
6&Teres Minor&Humerus&Scapula\\
7&Teres Major&Humerus&Scapula\\
8&Coracobra Chialis&Scapula&Humerus\\
\hline
\end{tabular}
\end{center}
\end{table}

\noindent The musculoskeletal system aims to mimic the movement of the shoulder. It was conceived as a single ball and socket joint. Hence, the three DOFs are allowed in a single point and are controlled by the eight selected muscles. The dynamic model of the system was developed following the DH convention (see Table \ref{table_Frames}) and its dynamics in joint coordinates are represented by the previous Eqn. \ref{DE1}, where $q^T= [q_1,q_2,q_3]$ is a vector of joint displacements, $q_1$ is the flexion displacement, $q_2$ is the inward rotation displacement and $q_3$ is the adduction displacement, $D(q)$ is the inertia matrix of the arm, $C(q,\dot{q})$ accounts for the centripetal and Coriolis effects, $g(q)$ is the gravity vector and $\tau_{muscles}$ are the torques generated by the muscle forces.

\begin{table}[ht]
\caption{Frames of the musculoskeletal system.}
\begin{center}
\label{table_Frames}
\begin{tabular}{ l l l l l l }
    \hline
    \multirow{2}{*}{Frame} &\multirow{2}{*}{Translation}& \multicolumn{4}{c}{DH} \\ \hhline{~~----}
 & &$\theta$&d&a&$\alpha$\\ \hline
Global&0&0 &0 & 0&0 \\ 
0&Humerus$_{(x;y;z)}$&0 &0 & 0&0 \\ 
1&0&q1 &$0$ & 0&-$\frac{\pi}{2}$    \\ 
2&0&q2 &$0$ & 0&0 \\ 
-&0&$\frac{\pi}{2}$ &$0$ & 0&$\frac{\pi}{2}$  \\
3&0&q3 &$0$ & 0&0 \\ 
\hline
\end{tabular}
\end{center}
\end{table}

\noindent The simulations are performed in order to move the arm to a static holding-a-cup position. The initial joint positions are randomly selected between $\pm$ $10^o$ from the desired final positions (see Table. \ref{table_pos_vel}).

\begin{table}[t]
\caption{Desired position, velocity and acceleration per DOF.}
\begin{center}
\label{table_pos_vel}
\begin{tabular}{ l l l l}
    \hline
   DOF&Target Position (DEG)& Initial Position (DEG)  \\ \hline
   1&$50^o$&$50^o\pm 10^o$\\
2&$27^o$&$27^o\pm 10^o$\\
3&$-45^o$&$-45^o\pm 10^o$\\
  \hline
\end{tabular}
\end{center}
\end{table}

\section{Results and Discussion}\label{S_Results}

\begin{figure}[ht]
\begin{center}
\includegraphics[width=3.34in]{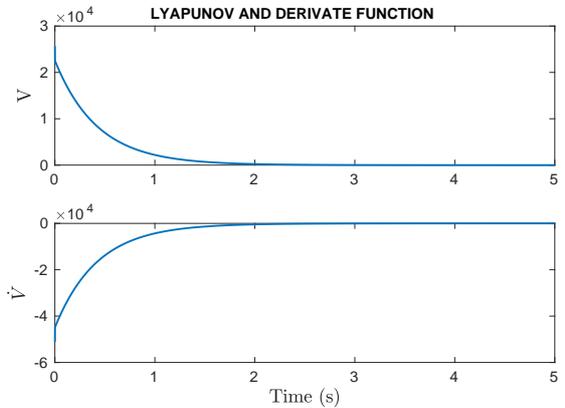}
\caption{Lyapunov and its derivative function.}
\label{fig_LYAP}
\end{center}
\end{figure}

\begin{figure}[ht]
\begin{center}
\includegraphics[width=3.34in]{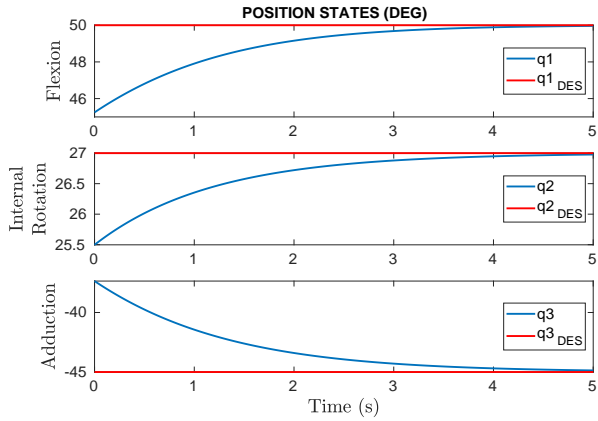}
\caption{Position tracking.}
\label{fig_POS_TRACK}
\end{center}
\end{figure}

\noindent The Lyapunov function and its time derivative can be seen on Fig. \ref{fig_LYAP}. Based on the positive definition of the Lyapunov function, the negative semi-definition of its time derivative and an expected convergence for both, the stability of the system is proved. Furthermore, from Fig. \ref{fig_POS_TRACK} can be seen that the states converge to the desired values in approximately 3.5 seconds. Consequently, the target position is successfully achieved by solving the muscle redundancy resolution.

\begin{figure}[ht]
\begin{center}
\includegraphics[width=3.34in]{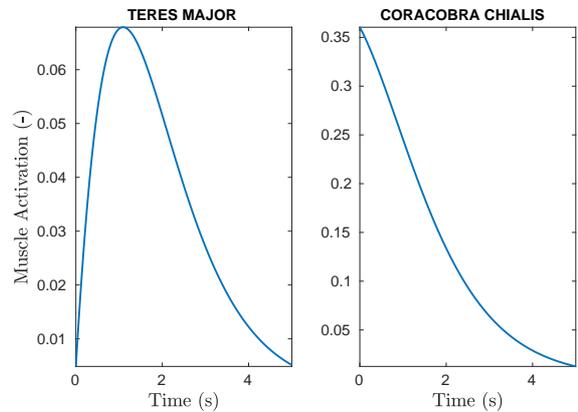}
\caption{Muscle activation of the Teres Major and Coracobra Chialis.}
\label{fig_MUSC7}
\end{center}
\end{figure}

\noindent The required muscle activations obtained to achieved the desired trajectory are shown in the following figures (Fig. \ref{fig_MUSC1}, \ref{fig_MUSC3}, \ref{fig_MUSC5} and \ref{fig_MUSC7}). Results show a convergence in each of the muscles. These converged values seem to be related to the required activations to overcome the gravity in the final position. The muscle activations show realistic values of magnitude and shape. However, it is difficult to quantify it without real-time experiments under the same experimental conditions. The muscles supraspinatus, infraspinatus, teres minor and coracobra chialis have shown the highest activation values. These highest activation makes sense due to the target position to be reached out.

\begin{figure}[ht]
\begin{center}
\includegraphics[width=3.34in]{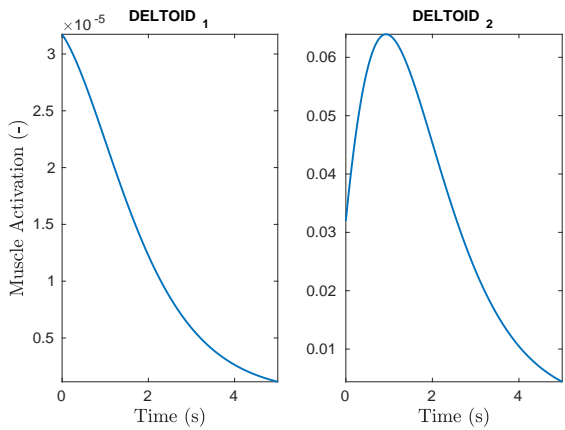}
\caption{Muscle activation of the Deltoid-1 and Deltoid-2.}
\label{fig_MUSC1}
\end{center}
\end{figure}

\begin{figure}[ht]
\begin{center}
\includegraphics[width=3.34in]{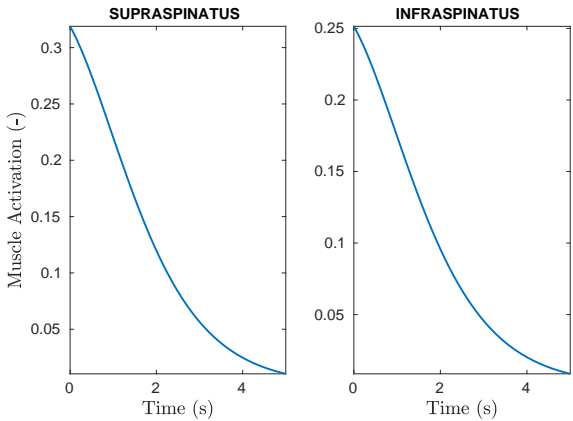}
\caption{Muscle activation of the Supraspinatus and Infraspinatus.}
\label{fig_MUSC3}
\end{center}
\end{figure}

\begin{figure}[ht]
\begin{center}
\includegraphics[width=3.34in]{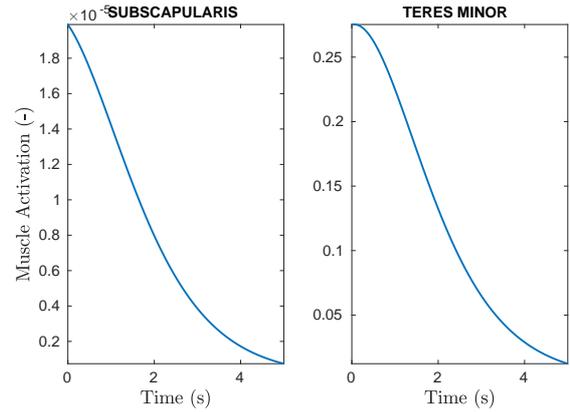}
\caption{Muscle activation of the Subscapularis and Teres Minor.}
\label{fig_MUSC5}
\end{center}
\end{figure}

\section{Conclusion and Future Work}\label{S_Future}

\noindent Upon completion of the feedback controller simulations on the shoulder model, realistic muscle activations were witnessed. Thus, the effectiveness of the feedback controller to solve muscle driven system with redundancy resolution is shown. It is important to consider that variations in the musculoskeletal distribution of the model, as well as a different dynamic models, can vary the results.

\noindent The feedback controller was successfully tested in simulation with the shoulder model. The objective of moving the arm from a static position to another one was achieved showing realistic results. Furthermore, results show a convergence on the muscle activations. This convergence suggests that the controller is effective finding the required muscle activation values to overcome the gravity effect on the musculoskeletal weight in the final position. Future research will be performed to compare these simulation results with real-time experiments with the same objectives.  

\noindent Several applications can make use of this method. Beyond the control tracking developed, the method can be adapted for different objectives. Some possible adaptations can be oriented to the optimization of human training or rehabilitation practices. For instance, advanced exercise machines could make use of the muscle activation as a feedback for performance improvement (through targeting or maximization of muscle activation). Likewise, rehabilitation developments could make use of the estimated muscle activation in order to provide a better assistance. Based on the muscle activation feedback, maximization of some muscle activation with minimization of others can be achieved.

\noindent In future developments, besides of more complex trajectories, more realistic and complete models will be used for testing the current approach. Further research will also include the integration of multiple control systems for human machine interaction related to advanced exercise machines and rehabilitation.

\section*{Acknowledgements}
\noindent This research was supported by the NSF, grant 1544702.

\bibliography{ifacconf}

\begin{thebibliography}{26}
\providecommand{\natexlab}[1]{#1}
\providecommand{\url}[1]{\texttt{#1}}
\providecommand{\urlprefix}{URL }
\expandafter\ifx\csname urlstyle\endcsname\relax
  \providecommand{\doi}[1]{doi:\discretionary{}{}{}#1}\else
  \providecommand{\doi}{doi:\discretionary{}{}{}\begingroup
  \urlstyle{rm}\Url}\fi

\bibitem[{A.~Hargens and Friden(1989)}]{ECC}
A.~Hargens, S.~Parazynski, M.A. and Friden, J. (1989).
\newblock Muscle changes with eccentric exercise: Implications on earth and in
  space.
\newblock Technical report, NASA.

\bibitem[{A.~Raptis and Valavanis(2011)}]{backstep_theo}
A.~Raptis, I. and Valavanis, K. (2011).
\newblock Linear and nonlinear control of small-scale unmanned helicopters.
\newblock 45.
\newblock \doi{10.1007/978-94-007-0023-9}.

\bibitem[{Bianchini et~al.(2019)Bianchini, Palmeri, Stefanini, Furbetta, and
  Di~Franco}]{Bianchini2019}
Bianchini, M., Palmeri, M., Stefanini, G., Furbetta, N., and Di~Franco, G.
  (2019).
\newblock The role of robotic-assisted surgery for the treatment of
  diverticular disease.
\newblock \emph{Journal of Robotic Surgery}.
\newblock \doi{10.1007/s11701-019-01008-y}.
\newblock \urlprefix\url{https://doi.org/10.1007/s11701-019-01008-y}.

\bibitem[{Chang and Kim(2013)}]{MOT2}
Chang, W.H. and Kim, Y.H. (2013).
\newblock Robot-assisted therapy in stroke rehabilitation.
\newblock \emph{Journal of Stroke}, 15, 174--181.

\bibitem[{{De las Casas} et~al.(2017){De las Casas}, Richter, and {v}an~den
  Bogert}]{MIPAPER1}
{De las Casas}, H., Richter, H., and {v}an~den Bogert, A. (2017).
\newblock Design and hybrid impedance control of a powered rowing machine.
\newblock In \emph{ASME 2017 Dynamic Systems and Control Conference}.

\bibitem[{{De las Casas}(2017)}]{HUM_THE}
{De las Casas}, H. (2017).
\newblock \emph{Design and Control of a Powered Rowing Machine with
  Programmable Impedance}.
\newblock Master's thesis, Cleveland State University.

\bibitem[{{De las Casas}(2019)}]{Humberto_GITHUB}
{De las Casas}, H. (2019).
\newblock Shoulder-model-3d-v1.
\newblock \url{https://github.com/HumbertoDlc/Shoulder-Model-3D-V1}.

\bibitem[{{De las Casas} et~al.(2019){De las Casas}, Kleis, Richter, Sparks,
  and Antonie{van den Bogert}}]{MIPAPER2}
{De las Casas}, H., Kleis, K., Richter, H., Sparks, K., and Antonie{van den
  Bogert} (2019).
\newblock Eccentric training with a powered rowing machine.
\newblock \emph{Medicine in Novel Technology and Devices}, 2, 100008.

\bibitem[{Delp et~al.(2007)Delp, Anderson, Arnold, Loan, Habib, John,
  Guendelman, and Thelen}]{OPENSIM}
Delp, S., Anderson, F., Arnold, A., Loan, P., Habib, A., John, C., Guendelman,
  E., and Thelen, D. (2007).
\newblock Opensim: Open-source software to create and analyze dynamic
  simulations of movement.
\newblock \emph{Biomedical Engineering, IEEE Transactions on}, 54, 1940 --
  1950.
\newblock \doi{10.1109/TBME.2007.901024}.

\bibitem[{Featherstone(2008)}]{IK}
Featherstone, R. (2008).
\newblock \emph{Inverse Dynamics}, 89--100.
\newblock Springer US, Boston, MA.
\newblock \doi{10.1007/978-1-4899-7560-7_5}.
\newblock \urlprefix\url{https://doi.org/10.1007/978-1-4899-7560-7_5}.

\bibitem[{Ghaoui(2005)}]{HMI_SIM}
Ghaoui, C. (2005).
\newblock \emph{Encyclopedia of Human Computer Interaction}.
\newblock Information Science Reference - Imprint of: IGI Publishing, Hershey,
  PA.

\bibitem[{Howe and Matsuoka(1999)}]{Robot_Surgery}
Howe, R.D. and Matsuoka, Y. (1999).
\newblock Robotics for surgery.
\newblock \emph{Annual Review of Biomedical Engineering}, 1(1), 211--240.
\newblock \doi{10.1146/annurev.bioeng.1.1.211}.
\newblock PMID: 11701488.

\bibitem[{Jung et~al.(2012)Jung, Park, Lee, and Kim}]{MOTOR}
Jung, D.W., Park, D.S., Lee, B.S., and Kim, M. (2012).
\newblock Development of a motor driven rowing machine with automatic
  functional electrical stimulation controller for individuals with paraplegia:
  a preliminary study.
\newblock In \emph{Annals of Rehabilitation Medicine}, 379--385.

\bibitem[{Khalaf et~al.(2015)Khalaf, Richter, {Van Den Bogert}, and
  Simon}]{POYA2}
Khalaf, P., Richter, H., {Van Den Bogert}, A., and Simon, D. (2015).
\newblock Multi-objective optimization of impedance parameters in a prosthesis
  test robot.
\newblock In \emph{ASME 2015 Dynamic Systems and Control Conference}, volume~3.

\bibitem[{Khalaf et~al.(2018)Khalaf, Warner, Hardin, Richter, and
  Simon}]{POYA1}
Khalaf, P., Warner, H., Hardin, E., Richter, H., and Simon, D. (2018).
\newblock Development and experimental validation of an energy regenerative
  prosthetic knee controller and prototype.
\newblock In \emph{ASME 2018 Dynamic Systems and Control Conference}, volume~1.

\bibitem[{Khummongkol and Yokota(2016)}]{HMI_Lang}
Khummongkol, R. and Yokota, M. (2016).
\newblock Computer simulation of human--robot interaction through natural
  language.
\newblock \emph{Artificial Life and Robotics}, 21(4), 510--519.

\bibitem[{Kistemaker et~al.(2013)Kistemaker, {Van Soest}, Wong, Kurtzer, and
  Gribble}]{HILL_MODEL}
Kistemaker, D., {Van Soest}, A., Wong, J., Kurtzer, I., and Gribble, P. (2013).
\newblock Control of position and movement is simplified by combined muscle
  spindle and golgi tendon organ feedback.
\newblock In \emph{Journal of Neurophysiology}.

\bibitem[{M.W.~Spong and Vidyasagar(2005)}]{ROBOT_MODEL}
M.W.~Spong, S.H. and Vidyasagar, M. (2005).
\newblock \emph{Robot Modeling and Control}.
\newblock Wiley.

\bibitem[{Ploutz-Snyder et~al.(2015)Ploutz-Snyder, Ryder, English, Haddad, and
  Baldwin}]{NASA_2}
Ploutz-Snyder, L., Ryder, J., English, K., Haddad, F., and Baldwin, K. (2015).
\newblock Risk of impaired performance due to reduced muscle mass, strength,
  and endurance.
\newblock Technical report, NASA.

\bibitem[{Rahman and Mizukawa(2013)}]{Sim_J1}
Rahman, M.A.A. and Mizukawa, M. (2013).
\newblock Model-based development and simulation for robotic systems with
  sysml, simulink and simscape profiles.
\newblock \emph{International Journal of Advanced Robotic Systems}, 10(2), 112.
\newblock \doi{10.5772/55533}.

\bibitem[{Richter and Warner(2017)}]{BACKSTEP_2D}
Richter, H. and Warner, H. (2017).
\newblock Backstepping control of a muscle-driven linkage.
\newblock In \emph{2017 IFAC World Congress}.

\bibitem[{Saul et~al.(2015)Saul, Hu, Goehler, Vidt, Daly, Velisar, and
  Murray}]{OPENSIM_MODEL}
Saul, K.R., Hu, X., Goehler, C.M., Vidt, M.E., Daly, M., Velisar, A., and
  Murray, W.M. (2015).
\newblock Benchmarking of dynamic simulation predictions in two software
  platforms using an upper limb musculoskeletal model.
\newblock In \emph{Computer Methods in Biomechanics and Biomedical
  Engineering}.

\bibitem[{Schmidt et~al.(2007)Schmidt, Werner, Bernhardt, Hesse, and
  Krüger}]{MOT3}
Schmidt, H., Werner, C., Bernhardt, R., Hesse, S., and Krüger, J. (2007).
\newblock Gait rehabilitation machines based on programmable footplates.
\newblock \emph{Journal of NeuroEngineering and Rehabilitation}.

\bibitem[{Sherman et~al.(2013)Sherman, Seth, and Delp}]{MOMENTARM}
Sherman, M., Seth, A., and Delp, S. (2013).
\newblock What is a moment arm? calculating muscle effectiveness in
  biomechanical models using generalized coordinates.
\newblock volume 2013.
\newblock \doi{10.1115/DETC2013-13633}.

\bibitem[{Tsumura(2008)}]{HMI_Hip}
Tsumura, H. (2008).
\newblock A computer simulation in surgery for a human hip joint.
\newblock \emph{Artificial Life and Robotics}, 12(1), 14--17.

\bibitem[{Vogt and Hoppeler(2014)}]{ECC2}
Vogt, M. and Hoppeler, H.H. (2014).
\newblock Eccentric exercise: mechanism and effects when used as training
  regime or training adjunct.
\newblock In \emph{The American Physiological Society}, 116, 1446--1454.

\end{thebibliography}



\end{document}